\documentclass[sigconf,screen]{acmart}
\renewcommand\footnotetextcopyrightpermission[1]{}
\usepackage{bibentry}
\usepackage{siunitx}
\usepackage{caption}
\usepackage{subcaption}
\usepackage{amsmath, amsthm}
\usepackage{mathtools}
\usepackage{booktabs}
\usepackage{enumitem}
\setdescription{style=unboxed, leftmargin=0cm, itemsep=0.8em}

\DeclareMathOperator*{\argmin}{arg\,min}

\AtBeginDocument{%
  \providecommand\BibTeX{{%
    \normalfont B\kern-0.5em{\scshape i\kern-0.25em b}\kern-0.8em\TeX}}}

\setcopyright{acmcopyright}
\copyrightyear{}
\acmYear{}
\acmDOI{}

\acmConference[Seminar in Data Science '19]{Seminar in Data Science}{2019}{Linz, Austria}
\acmBooktitle{}
\acmPrice{}
\acmISBN{}

\begin{document}

\title{Graph Neural Networks for Node-Level Predictions}

\author{Christoph Heindl}
\email{christoph.heindl@gmail.com}
\orcid{0000-0002-6362-8976}
\affiliation{%
  \institution{JKU - Institute of Computational Perception}
  \city{Linz}
  \country{Austria}
  \vspace{1cm}
}

\vspace{1cm}

\begin{abstract}
  The success of deep learning has revolutionized many fields of research including areas of computer vision, text and speech processing. Enormous research efforts have led to numerous methods that are capable of efficiently analyzing data, especially in the Euclidean space. However, many problems are posed in non-Euclidean domains modeled as general graphs with complex connection patterns. Increased problem complexity and computational power constraints have limited early approaches to static and small-sized graphs. In recent years, a rising interest in machine learning on graph-structured data has been accompanied by improved methods that overcome the limitations of their predecessors. These methods paved the way for dealing with large-scale and time-dynamic graphs. This work aims to provide an overview of early and modern graph neural network based machine learning methods for node-level prediction tasks. Under the umbrella of taxonomies already established in the literature, we explain the core concepts and provide detailed explanations for convolutional methods that have had strong impact. In addition, we introduce common benchmarks and present selected applications from various areas. Finally, we discuss open problems for further research.
\end{abstract}

\maketitle

\setlength{\parindent}{0em}
\setlength{\parskip}{0.8em}

\section{Introduction}
The end of the last millennium marks the beginning of a revolution in machine learning. Gradient-based learning applied to multi-layer neural networks showed that feature representations can be learned instead of creating them manually \cite{lecun1998}. In the following years, increased computing power led to network architectures with an increasing number of hidden layers. Such deeply learned architectures often outperformed conventional methods by large margins. Successful examples include image classification \cite{krizhevsky2012imagenet}, speech sentence recognition \cite{dahl2011context} and text translation \cite{gehring2017convolutional}. The success of the aforementioned examples is to a large extend based on two properties: a) an Euclidean data domain is underlying the regular structure of images, language and text and b) convolutional filters are capable of efficiently processing data in such Euclidean domains. 

\begin{figure}[tp]
  \centering
  \begin{subfigure}[b]{0.48\columnwidth}
    \centering
    \includegraphics[width=0.8\hsize]{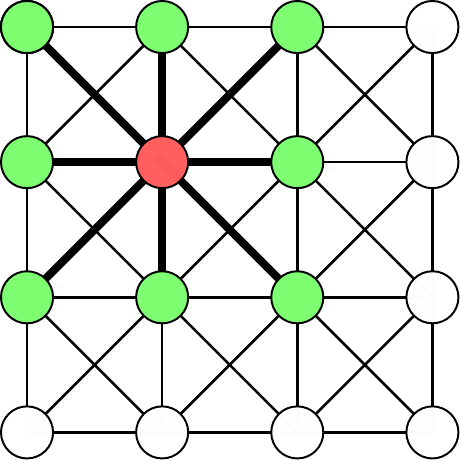}
    \caption{\label{fig:regular} Euclidean domain. Green nodes are 2D convolutional neighbors of red node in regular grid. Topology of neighborhood is constant.}
  \end{subfigure}
  \hfill
  \begin{subfigure}[b]{0.48\columnwidth}
    \centering
    \includegraphics[width=\hsize]{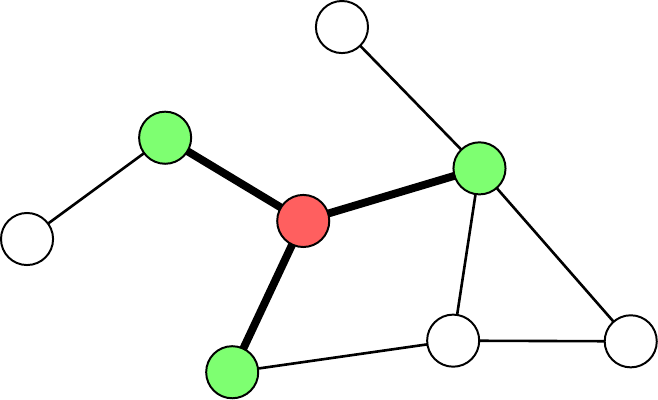}
    \caption{\label{fig:graph} Non-Euclidean domain. Green nodes are convolutional neighbors of red node. Topology of neighborhood is irregular.}
  \end{subfigure}
  \caption{\label{fig:regularvsgraph} Data structure in Euclidean and non-Euclidean domains.}
  \vspace{-0.2cm}
\end{figure}

In recent years the interest in applying machine learning to non-Euclidean domains has increased. Non-Euclidean data does not exhibit a regular structure such as images/sound or text, but instead is modelled by arbitrary graphs consisting of nodes, edges and attributes. Applications of this form are widespread \cite{bronstein2017geometric}: In computer graphics operations on 3D objects can be represented as methods operating on mesh-graphs. In social networks the characteristics of users can be modelled as properties/signals on top of a social-graph. In sensor networks, the collection of sensors can be seen as elements of an interconnected network-graph. Motivated by the wide range of possible applications, research focused on generalizing deep learning architectures to handle the complexity of graph data. Specifically, many modern Graph Neural Networks (GNNs) are based on a generalization of Euclidean convolution operation. Figure~\ref{fig:regularvsgraph} indicates how graph convolutions compare to Euclidean convolutions.

This paper attempts to be a gentle introduction into the development of GNNs and their applications. In particular, we present Convolutional Graph Neural Networks (ConvGNNs) in depth, which attempt to generalize convolution from the Euclidean to the non-Euclidean domain. Due to the vast number of methods published in recent years, this work focuses on few methods which are elaborated in more detail. For a more comprehensive summary of other methods see \cite{wu2019comprehensive, zhou2018graph}. 

This work is structured as follows. We start by introducing GNNs from a general perspective in Section~\ref{sec:gnn}. The abstract view is used to familiarize the reader with an overview of learning and training methods in Section~\ref{sec:learning}. Starting with Section~\ref{sec:recgnn}, pioneering works in the field of GNNs are highlighted. Section~\ref{sec:convgnn}, the main section of this work, deals with ConvGNNs in particular. In Section~\ref{sec:benchmarks} we compare the presented methods against standard datasets and in Section~\ref{sec:applications} we show successful applications of GNNs for open problems in medicine, physics and social sciences. Section~\ref{sec:discussion} discusses open problems and future research. Finally, we conclude this work in Section~\ref{sec:conclusion}.

\section{Graph Neural Networks}
\label{sec:gnn}
A Graph Neural Network (GNN) can be seen as a neural network operating directly on a graph structure. This section introduces GNNs from an abstract view point.

\subsection{Notation}
Throughout this work we use lower-case non-bold characters $x$ to denote scalars or scalar functions. Bold-faced lower-case characters $\mathbf{x}$ represent column vectors and upper-case bold characters $\mathbf{A}$ matrices. $\mathbf{x}_i$ denotes the $i$-th element of $x$, $\mathbf{A}_{ij}$ the element at the $i$-th row and $j$-th column of $\mathbf{A}$. We use $\mathbf{A}_{:,i}$ and $\mathbf{A}_{i,:}$ to denote the $i$-th column and row of $\mathbf{A}$. Superscripts with lower case letters $^t$ are used to indicate a time/layer indices and $^T$, $^-1$ denote matrix transpose and inverse. Other commonly used symbols are listed in Table~\ref{tab:symbols}.

\subsection{Undirected Graph}
Although many specific forms of graphs exist, this work's scope is restricted to undirected graphs, which we define as follows. An undirected graph $G = (\mathcal{V}, \mathcal{E})$ is a collection of nodes $\mathcal{V}$ and edges $\mathcal{E}$. Two nodes $v_i, v_j \in \mathcal{V}$ are said to be connected if $e_{ij}=(v_i, v_j) \in \mathcal{E}$. If $e_{ij} \in \mathcal{E}$ then also $e_{ji} \in \mathcal{E}$. In addition, let $v$ be the $i$-th node, then set of characteristics associated with with $v$ is the $i$-th row of feature matrix $\mathbf{X} \in \mathbb{R}^{N \times D}$, denoted by $\mathbf{X}_{i,:}$.

\subsection{Adjacency Matrices}
Let $G$ be an undirected graph and let $|\mathcal{V}|=N$, then $\mathbf{A}$ denotes the binary $N \times N$ adjacency matrix defined to be 
\begin{equation}
  \mathbf{A}_{ij} = \left\{
    \begin{array}{ll}
      1 & \textrm{if}\quad e_{ij} \in \mathcal{E} \\
      0 & \textrm{otherwise}
    \end{array}
  \right\}.
\end{equation}
For undirected graphs $\mathbf{A}$ is symmetric, that is $\mathbf{A} = \mathbf{A}^T$. 

\subsubsection{Self-Connectivity}
Unless otherwise stated, $\mathbf{A}_{ii}$ is zero. That is, the nodes of $G$ are not self-connected. In GNNs, however, it often makes sense to include self-connections to enable compact formulations in the context of feature transformation. Therefore, we denote by $\bar{\mathbf{A}}$ the adjacency matrix with self-connections given by
\begin{equation}
  \bar{\mathbf{A}} = \mathbf{I} + \mathbf{A},
\end{equation}
where $\mathbf{I} \in \mathbb{R}^{N \times N}$ is the identity matrix.

\subsubsection{Weighted Adjacency}
So far, only binary $\{0,1\}$ adjacency matrices have been presented. However, it is useful to think about arbitrary real-valued entries of $\bar{\mathbf{A}}$ to encode weighted connectivity. For example, we can model a set of $N$ unstructured points in $\mathbb{R}^D$ by inducing the following graph topology: Assume the coordinates of all points is given by the feature matrix $\mathbf{X} \in \mathbb{R}^{N \times D}$. Then, one way to define a weighted adjacency matrix is as follows
\begin{equation}
  \bar{\mathbf{A}}_{ij} = \exp\left(-\frac{d(\mathbf{X}_{i,:}, \mathbf{X}_{j,:})}{\sigma^2}\right),
\end{equation}
where $d(\cdot, \cdot)$ is a metric on the Euclidean space and $\sigma$ is a scaling factor. Note, that the induced weighted adjacency matrix is self-connected, because $d(\mathbf{X}_{i,:}, \mathbf{X}_{i,:}) = 0$. Such a weighted adjacency matrix densely connects all points, weighted by an exponential fast decaying value based on their geometric distance.

For the purpose of this work, all adjacency matrices (unless otherwise noted) will refer to a simpler, binary adjacency matrix.

\subsubsection{Normalization}
Given $\bar{\mathbf{A}}$, it insightful to think about the following linear action
\begin{equation}
  \mathbf{Z} = \bar{\mathbf{A}}\mathbf{X}, \label{eq:transformfeature}
\end{equation}
where $\mathbf{X}$ is the feature input matrix and $\mathbf{Z}$ is the transformed feature matrix. For a single node $i$, the action can be written as follows
\begin{equation}
  \mathbf{Z}_{i,j} = \sum_{k=1}^{N} \bar{\mathbf{A}}_{i,k}\mathbf{X}_{k,j}.
\end{equation}
Which says that the $i$-th transformed feature in $j$-th coordinate is a weighted linear combination of the input features in $j$-th coordinated of neighbors of node $i$. In case of a binary adjacency matrix, the above becomes a simple sum. In GNNs, Equation~\ref{eq:transformfeature} may appear in an recursive fashion, by either layer stacking or recurrent iteration. Because $\bar{\mathbf{A}}$ is not normalized, the linear actions scale the feature values in an undesired fashion. Additionally, nodes connected to many other nodes will have the same importance as nodes with fewer neighbors. 

To fix this, we apply adjacency matrix normalization. Let $\mathbf{D}$ denote the diagonal node degree matrix $\mathbf{D}_{ii} = \sum_{j=1}^{N}\bar{\mathbf{A}}_{ij}$. The following normalization variants are commonly found
\begin{enumerate}
  \item Row normalization $\hat{\mathbf{A}} = \mathbf{D}^{-1}\bar{\mathbf{A}}$ corresponds to $$\mathbf{Z}_{i,j} = \frac{1}{\mathbf{D}_{ii}}\sum_{k=1}^{N} \bar{\mathbf{A}}_{i,k}\mathbf{X}_{k,j},$$ i.e computing average of neighboring features.
  \item Column normalization $\hat{\mathbf{A}} = \bar{\mathbf{A}}\mathbf{D}^{-1}$ corresponds to $$\mathbf{Z}_{i,j} = \sum_{k=1}^{N} \frac{\bar{\mathbf{A}}_{i,k}\mathbf{X}_{k,j}}{\mathbf{D}_{kk}},$$ which effectively sums over neighboring features normalized by the number of their neighbors.
  \item Symmetric normalization can be done naively $\hat{\mathbf{A}} = \mathbf{D}^{-1}\bar{\mathbf{A}}\mathbf{D}^{-1}$, which would lead to vanishing feature values when iterating long enough due to the denominator $$\mathbf{Z}_{i,j} = \sum_{k=1}^{N} \frac{\bar{\mathbf{A}}_{i,k}\mathbf{X}_{k,j}}{\mathbf{D}_{ii}\mathbf{D}_{kk}},$$ which is why it generally not used. Instead, better dynamics are achieved by \begin{equation}
    \hat{\mathbf{A}} = \mathbf{D}^{-0.5}\bar{\mathbf{A}}\mathbf{D}^{-0.5}, \label{eq:normalization}
  \end{equation} which corresponds to 
  $$\mathbf{Z}_{i,j} = \sum_{k=1}^{N} \frac{\bar{\mathbf{A}}_{i,k}\mathbf{X}_{k,j}}{\sqrt{\mathbf{D}_{ii}}\sqrt{\mathbf{D}_{kk}}}.$$ 
\end{enumerate}

Unless otherwise stated, this paper refers to Equation~\ref{eq:normalization} when referring to normalized adjacency $\hat{\mathbf{A}}$ matrices. One drawback of normalization is that it cannot handle isolated vertices, as $\mathbf{D}^{-1}$ is not defined in such cases. A quick fix seen in implementations is to add a small constant to the entries of $\mathbf{D}$.

\begin{table}[htp]
  \centering        
  \begin{tabular}{cl}
  \toprule
  Symbol & Meaning \\ 
  \midrule
  $G(\mathcal{V}, \mathcal{E})$ & Undirected graph with nodes $\mathcal{V}$ and edges $\mathcal{E}$ \\
  $N$ & Number of nodes $|\mathcal{V}|$ \\
  $D$ & Number of input dimensions \\
  $K$ & Number of output dimensions \\
  $\mathbf{X}, \mathbf{Y}, \mathbf{Z}$ & Input/output feature matrices\\
  $\mathbf{x}$ & column vector\\
  $\mathbf{I}$ & Identity matrix \\
  $\mathbf{W}$ & Parameter matrix \\
  $\mathbf{D}$ & Diagonal node degree matrix \\
  $\mathbf{A}$ & binary/weighted Adjacency matrix \\
  $\bar{\mathbf{A}}$ & Adjacency matrix with self-loops \\
  $\hat{\mathbf{A}}$ & Symmetrically normalized adjacency matrix \\
  $\mathbf{L}$ & Graph Laplacian \\
  $\hat{\mathbf{L}}$ & Symmetrically normalized Graph Laplacian \\
  $\sigma$ & Element-wise activation function \\
  \bottomrule
  \end{tabular}
  \caption{\label{tab:symbols} Common symbols and their meaning used throughout this work.}
\end{table}

\subsection{Network Outputs}
Like any other neural network, a GNN can be seen as a computational graph assembled from re-usable building blocks, which we usually call layers. In contrast to conventional neural networks, these layers operate on graphs or transformations thereof. In the context of this work, transformation refers to either changes in the topology of the graph, associated node features or both. As such, GNNs can serve a variety of prediction purposes. We categorize GNNs based on their analytic purpose as follows \cite{wu2019comprehensive}:
\begin{description}
  \item[Node-level] A GNN operating on node-level computes values for each node in the graph and is thus useful for node classification and regression purposes. In addition such GNNs can be seen as building blocks for computing hidden node embeddings.
  \item[Edge-level] These type of GNNs are used to predict values for each graph edge, or a transformed version of it.
  \item[Graph-level] Refers to GNNs that predict a single value for an entire graph. Mostly used for classifying entire graphs or computing similarities between graphs.
\end{description}

This work is mainly concerned with node-level outputs. As we will see shortly, node-level outputs are re-useable units for all other analytic tasks. 

\subsection{Node-level GNNs}
Our definition of a graph neural network is restricted to node-level tasks and defined to be a function $g$ of the following form
\begin{equation}
  \hat{\mathbf{Y}} = g(G, \mathbf{X}; \Omega), \label{eq:gnn}
\end{equation} 
where $\Omega$ is a set of trainable network parameters, $\hat{\mathbf{Y}} \in \mathbb{R}^{N \times K}$ summarizes the model's prediction for each node, $G$ is the graph topology as defined before and $\mathbf{X}$ represents node feature vectors.

\section{Learning \& Training}
\label{sec:learning}

\subsection{Learning Variants}
For GNNs and in machine learning in general, we distinguish two main approaches to learning from data \cite{vapnik1998statistical}:
\begin{description}
  \item[Inductive] Induction refers to the task of learning patterns from data such that the model generalizes to any new, unseen data points. That is, we can use the model as a surrogate for the unknown true function.
  \item[Transductive] In transductive learning we relax the idea of building a model that generalizes to any new data points. Instead, one attempts to predict values for examples already known during training. If new samples are provided, the learning algorithm might need to be applied again.
\end{description}
Take the task of function approximation as an example. In inductive learning one attempts to learn a model that we can use to evaluate the target function at any point. In contrary, transductive learning computes function values for specific locations already known during the training process and avoids learning universally applicable model.

\subsection{Training Objectives}
In addition to the learning process, we can distinguish methods for training a GNN with node-level outputs. GNNs are trained like other neural networks by (stochastic) gradient descent of an objective function with respect to parameters. Given a (supervised) training set $\{\mathbf{X},\mathbf{Y}\}$ on the graph G, we can distinguish the following methods.
\begin{description}
  \item[Supervised] The goal of supervised training is to minimize 
  \begin{equation}
    \Omega^* = \argmin_\Omega \sum_{i=1}^N L(\mathbf{Y}_{i,:}, \hat{\mathbf{Y}}_{i,:}),
  \end{equation} 
  where $L$ is node-level loss function, and $\hat{\mathbf{Y}}_{i,:}$ is the i-th row of $g(G, \mathbf{X}; \Omega)$.
  \item[Semi-supervised] Semi-supervised training takes unlabelled data into account. This is done via constructing topology based loss functions that act like smoothing/regularization constraints. In \cite{yang2016revisiting} a penalty term based on the dissimilarity of predictions for neighboring graph nodes is used
  $$
    L_{reg} = 0.5\sum_{i,j} \mathbf{A}_{ij}\,d(\hat{\mathbf{Y}}_{i,:}, \hat{\mathbf{Y}}_{j,:}),
  $$
  where $\mathbf{A}$ is the adjacency matrix and $d$ is a distance metric. Combined with the supervised loss, the semi-supervised target becomes
  $$
    \Omega^* = \argmin_\Omega \left[ \sum_{i=1}^N L(\mathbf{Y}_{i,:}, \hat{\mathbf{Y}}_{i,:}) + \lambda L_{reg} \right],
  $$
  where $\lambda$ is balancing factor.
  \item[Unsupervised] In this scenario, no supervised training data is used. The loss functions in this scenario enforce similar node-embeddings of nearby graph elements \cite{hamilton2017inductive} or use reconstruction errors \cite{kipf2016variational}.  
\end{description}

Here we focus on inductive learning with either supervised or semi-supervised training objectives. Keep in mind that, in the above description, we made the implicit assumption that the graph topology remains constant for each training sample. In practice this is often not the case and training needs to extend to datasets of the form $\{(\mathbf{G}_i, \mathbf{X}_i, \mathbf{Y}_i)\}_{i \le T}$. Unless otherwise stated, we assume a single constant graph topology throughout this treatment.

\subsubsection{Batch vs. Stochastic Training}
Up until recently, GNN training was performed primarily in batch mode. That is, network forward and backward passes were performed for all graph nodes at once (usually including train and test nodes). As graphs grew in size, methods for mini-batch training became valuable. On graphs, mini-batch training poses a problem: each node is connected to variable number of neighbors and this causes issues with modern training architectures, which use tensors as their building block to store information. 

One solution is given in \cite{hamilton2017inductive}. They propose to perform random sampling in local neighborhoods for each layer to keep the mini-batch size constant. The sampling strategy keeps the mini-batch size constant throughout many network layers and introduces a sort of stochasticity, often desired in stochastic optimization.

\section{Recurrent Graph Neural Networks}
\label{sec:recgnn}
Recurrent Graph Neural Networks (RecGNNs) \cite{sperduti1997supervised, scarselli2008graph} are among the early works in the field of graph neural networks. These methods iteratively apply the same parametrized function to node values to extract high level information patterns. The information is propagated via the edges of the graph until an equilibrium is reached. Conceptually this is reminiscent of methods for inferring marginal probabilities in graphical models, such as (loopy) belief propagation \cite{pearl1986probabilistic}. 

The iterative update rule for a RecGNN can be defined as follows
\begin{equation}
  \mathbf{H}^t = f(G, \mathbf{H}^{t-1}; \mathbf{W})
\end{equation}
with $\mathbf{H}^t \in \mathbb{R}^{N \times D}$ being the feature representation at level $t$, $\mathbf{H}^0 = \mathbf{X}$, and $f$ being a function parametrized by $\mathbf{W} \in \mathbb{R}^{D \times D}$. Here $\mathbf{H}^t$ can be seen as a hidden state after the $t$-th iteration. The output of a RecGNN is simply the hidden state after the last iteration $T$
\begin{equation}
  \hat{\mathbf{Y}} = g(G, \mathbf{X}; \Omega) = \mathbf{H}^T,  
\end{equation}
with $\Omega = \{\mathbf{W}\}$. Since parameters are shared across iterations, the dimensionality of input and hidden states does not change. In order for the above update procedure to converge, $f$ needs to shrink the distance of hidden states. That is, $f$ needs to be contraction mapping. 

The following update rule specifies a typical RecGNN
\begin{equation}
  \mathbf{H}^t_{i,:} = \sigma\left(\mathbf{H}^{t-1}_{i,:}\mathbf{W}  + \sum_{n \in \mathcal{N}_G(i)} \mathbf{H}^{t-1}_{n,:}\mathbf{W} \right),
\end{equation}
where $\mathcal{N}_G(i)$ is the set of nodes connected to the i-th node via $G$, and $\sigma$ is a non-linear function applied element-wise. The above update rule computes new hidden state as a non-linear function acting on the sum of transformed previous hidden states of that node and it's neighbors. We can write this more compactly for all of $G$'s nodes as
\begin{equation}
  \mathbf{H}^t = \sigma\left(\hat{\mathbf{A}}\mathbf{H}^{t-1}\mathbf{W}\right),
\end{equation}
where $\hat{\mathbf{A}}$ is the symmetrically normalized adjacency matrix of $G$ extended by self loops. Observe that normalization of the adjacency matrix may have been skipped in source literature.

\section{Convolutional Graph Neural Networks}
\label{sec:convgnn}
Convolutional Graph Neural Networks (ConvGNNs) attempt to generalize convolutions from the Euclidean domain to graph structures. Similar to RecGNNs, ConvGNNs compute hidden states by aggregation of neighboring hidden states. What distinguishes them from RecGNNs is that ConvGNNs stack multiple layers of graph convolutions to extract high level node information. As such, they replace iteration with a fixed number of layers carrying different parameter sets. In the following, we first highlight the main properties of Euclidean convolutions and then explain two branches of {ConvGNNs}: Spectral ConvGNNs heavily rely on graph signal processing \cite{chung1997spectral} and the Graph Fourier transform to define graph signal filters. The filters of spectral GNNs can be seen as global filters. In contrast spatial ConvGNNs propose filters operating on the local neighborhood of graphs directly. Therefore, they are usually more scalable for larger graphs and are also suitable for tasks with different graph topologies in inductive learning situations (see Section~\ref{sec:learning}).

\subsection{Properties of Euclidean Convolutions}
\label{sec:euclideanconv}
\begin{figure}[htp]
  \begin{subfigure}[b]{0.40\columnwidth}
    \includegraphics[width=\hsize]{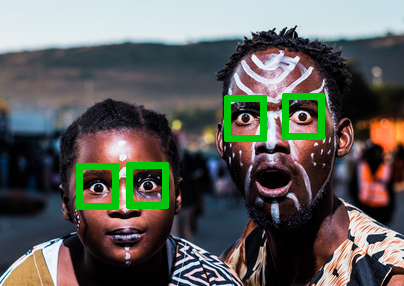}
    \caption{\label{fig:locstat} Locality and Stationarity. Features are translational invariant and compact.}
  \end{subfigure}
  \hfill
  \begin{subfigure}[b]{0.56\columnwidth}
    \includegraphics[width=0.98\hsize]{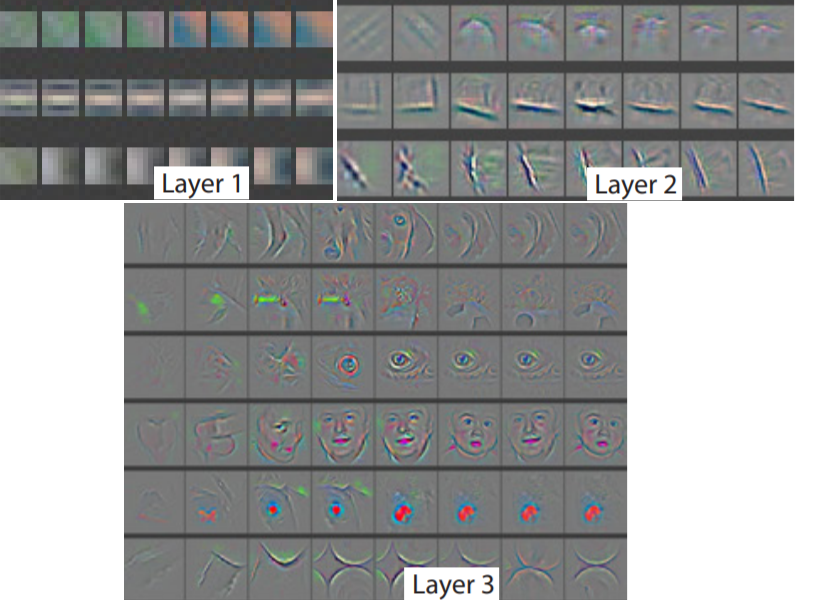}
    \caption{\label{fig:multiscale} Multi-scale. Complex feature aggregation through hierarchical composition \cite{zeiler2014visualizing}.}
  \end{subfigure}
  \caption{\label{fig:features} Compositional feature principles for 2D images.}
\end{figure}

Images and video can be considered functions on the Euclidean space, sampled on a grid-like regular topology (see Figure~\ref{fig:regular}). Similarly, sound can be modelled by amplitude as a function of time, sampled on a one-dimensional time-line. Albeit such data is extremely high-dimensional, convolution neural networks \cite{fukushima1980neocognitron} have proven powerful tools to efficiently and robustly process it. The basic underlying assumption is that audio/video/text is \emph{compositional} \cite{henaff2015, bronstein2017geometric} in nature and thus can be efficiently processed by filters that adhere to the following properties:
\begin{description}
  \item[Locality] Features can be described by a local compact receptive field. 
  \item[Stationarity] Features are independent of their location in the domain, i.e. translational invariant.  
  \item[Multi-scale] Complex features can be build from simpler ones through aggregation in hierarchies.
\end{description}
Convolutional filters exploit these three principles as follows. Locality is ensured by a compact fixed-size filter mask. Translational invariance is achieved by sliding the filter mask across the sampled Euclidean grid. Finally, complex features are created by applying convolutions to down-sampled features created by convolutions operating on higher resolution input. The three properties are illustrated in Figure~\ref{fig:features}. 

Mathematically, such convolutions can be compactly expressed by the cross-correlation operator  $\star$ applied to two real-valued functions $f$ and $h$
\begin{equation}
  (f \star h)(x) = \int_{-\infty}^{\infty} f(t)h(t+x)\,dt,
\end{equation}
where we ignored a bias term. In the discrete case, this amounts to shifting the compact filter $h$ across $f$ and computing point-wise inner products. The inner product acts as an \emph{aggregator} that condenses the contents of the input with the filter. Because of the regular underlying structure, Euclidean convolution exhibit the following two important properties:
\begin{enumerate} 
  \item The inner product always comprises the same number of elements.
  \item The elements of the inner product are always processed in the same order, giving filters a sense of orientation.
\end{enumerate}
As we will see below, it is challenging to generalize these properties to arbitrary graphs.

From an implementation perspective, convolution is an efficient method for processing Euclidean data: Locality is bounded by the number of parameters which is constant $\mathcal{O}(1)$, stationarity for compact kernels requires $\mathcal{O}(n)$ operations or $\mathcal{O}(n \log n)$ for general Fourier transform. Finally, generating multi-scale hierarchies (i.e. pooling, down-sampling) is bounded by $\mathcal{O}(n)$.

\subsection{Spectral ConvGNNs}
Spectral ConvGNNs rely on the Graph Fourier Transform to perform signal processing on graphs. A graph signal $$x \colon \mathcal{V} \rightarrow \mathbb{R}$$ is mapping from the vertices of a graph to real numbers. Let ${\mathbf{x} \in \mathbb{R}^N}$ be the column-vector representation of a signal, that is the $i$-th element of $\mathbf{x}$ corresponds to the signal value of node $i$. The node feature vectors $\mathbf{X}$ can thus be seen as series of $K$ independent signals. Graph processing seeks for a new basis to decompose the signal into a set of mutually orthogonal components. Central for this operation is the Graph Laplacian \cite{chung1997spectral}
\begin{equation}
  \mathbf{L} = \mathbf{D} - \mathbf{A},
\end{equation}
where $\mathbf{D}$ is again the diagonal degree matrix. As with adjacency matrices, in practice we rather work with symmetrically normalized Graph Laplacians
\begin{equation}
  \hat{\mathbf{L}} = \mathbf{I} - \hat{\mathbf{A}}.
\end{equation} The Graph Laplacian is the discrete form of the Laplacian $$\Delta x = \nabla^2 x,$$ measuring the smoothness of graph signals. The key idea is that a smooth graph signal does not change its value by much from one vertex to another connected vertex. The Graph Laplacian is a real symmetric positive semidefinite matrix and can thus be factored as $$\hat{\mathbf{L}} = \mathbf{U}\mathbf{V}\mathbf{U}^T,$$ where $\mathbf{V}$ is a diagonal ascendingly sorted matrix of eigenvalues, and $\mathbf{U}$ is the corresponding matrix of eigenvectors. The Graph Fourier Transform of a signal $x$ is defined to be the projection of $x$ onto the basis induced by $\mathbf{U}$
\begin{equation}
  \hat{\mathbf{x}} = \mathbf{U}^T\mathbf{x}.
\end{equation}
Note, that in practice often only the first few eigenvectors are used to capture the graph smoothness. The Graph Convolution $\star_G$ of two signals $x$, $w$ in their corresponding vector representation is then defined as $$x \star_G w = \mathbf{U}(\mathbf{U}^T\mathbf{x} \odot \mathbf{U}^T\mathbf{w}),$$ where $\odot$ represents element-wise multiplication. Let
\begin{equation}
  \mathbf{W} = diag(\mathbf{U}^T\mathbf{w}) \label{eq:diagw}
\end{equation}
denote a non-parametric filter (i.e all $N$ parameters are free) in $\mathbb{R}^{N \times N}$, then the Graph Convolution can be compactly written as $$x \star_G w = \mathbf{U}\mathbf{W}\mathbf{U}^T\mathbf{x}.$$ 

In \cite{bruna13spectral} Spectral Convolutional Neural Networks (SpectralCNN) are presented. The learnable parameters per layer are the diagonal entries of $\mathbf{W}$ in Equation~\ref{eq:diagw}. 
The output signal $j$ in the $t$-th layer of a SpectralCNN is computed via
\begin{equation}
  \mathbf{H}^t_{:,j} = \sigma\left(\mathbf{U} \sum_{i=1}^{D_t} \prescript{j}{i}{\mathbf{W}^t} \mathbf{U}^T \mathbf{H}^{t-1}_{:, i} \right),
\end{equation}
where $\prescript{j}{i}{\mathbf{W}^t} \in \mathbb{R}^{N \times N}$ is the diagonal parameter matrix in layer $t$ transforming the input signal $i$ of layer $t-1$ to output signal $j$, and $D_t$ is the number of input signals in $\mathbf{H}^{t-1}$. 

The spectral filters presented so far violate Euclidean convolution properties in the following sense: the learned filters are not localized in space, as each filter is composed of $N = |\mathcal{V}|$ parameters. In addition, a perturbation of vertex ordering changes the eigenvector basis which makes learned filters domain specific, i.e cannot be applied in a different context. The performance of the method is further limited by the requirement of performing an eigen-decomposition $\mathcal{O}(N^3)$. For this reason various improvements to the SpectralCNN architecture have been proposed. For example, in \cite{defferrard2016convolutional} parametric filters based on recursive Chebyshev polynomials up to order $K$ are introduced. In their method, polynomic coefficients are shared across graph locations to ensure feature locality. Additionally, the proposed method offers linear computational complexity and constant learning complexity (depending on the order of $K$).

\subsection{Spatial ConvGNNs}
Spatial ConvGNNs define filters directly on the graph neighborhood, by stacking non-linear aggregation functions defined on the local neighborhood of nodes. Since a graph topology generally does not define an explicit neighboring order, these aggregation functions need to be permutation invariant. As such, spatial graph convolutions lack orientation as described in Section~\ref{sec:euclideanconv}. Learnable filters correspond to symmetric variants of Euclidean kernels. Compared to spectral approaches, spatial ConvGNNs avoid global processing, making this method scalable even for large graphs. In addition, spatial approaches are suitable for learning inductive models that may be applied even to altering graph topologies. RecGNNs and spatial ConvGNNs share similar ideas, but what sets them apart is the fact that spatial methods eliminate weight sharing and replace iterative processing by a fixed number of stacked layers.

\cite{micheli2009neural} proposes the following layer-wise architecture
\begin{equation}
  \mathbf{H}^{t} = \sigma\left(\mathbf{X}\mathbf{W}^t + \sum_{k=1}^{t-1} \mathbf{A}\mathbf{H}^{k}\prescript{}{k}{\mathbf{W}^t} \right), \label{eq:nn4g}
\end{equation}
where $\prescript{}{k}{\mathbf{W}^t}$ is a parameter matrix transforming features $\mathbf{H}^{k}$ of layer $k$ to layer $t$. The model, coined NN4G, offers skip-connections from all previous layers to the current layer. NN4G is trained layer-wise, and was used for graph-level predictions. 

\cite{kipf2017graph} recently proposed a spatial graph convolution based on first-order approximation of spectral methods
\begin{equation}
  \mathbf{H}^{t} = \sigma\left(\hat{\mathbf{A}}\mathbf{H}^{t-1}\mathbf{W}^t \right), \label{eq:gcn}
\end{equation}
with $\mathbf{H}^0 = \mathbf{X}$. Equation~\ref{eq:gcn} is similar to Equation~\ref{eq:nn4g} without skip-connections and normalization.

\cite{hamilton2017inductive} refines Equation~\ref{eq:gcn} by separating functions for aggregation and stacking. While in  Equation~\ref{eq:gcn} all features in the convolutional radius are treated the same, Hamilton et al. investigate combinations of neighborhood aggregation (sum, mean, LSTM/GRU) followed by concatenation with center node features. Further spatial methods are detailed in \cite{wu2019comprehensive} and \cite{zhou2018graph}.

\subsection{Beyond ConvGNNs}
ConvGNNs introduced in the previous sections play an important role as building blocks for more complex architectures.

\subsubsection{Graph Attention Networks}
Graph attention networks introduced by \cite{velickovic2018graph} extend constant adjacency matrices by per-node attention weights. Using a shallow neural network, each node attends over its neighboring nodes in the following way $$\mathbf{A}^t_{ij} = f(\mathbf{H}^{t-1}_{i,:}, \mathbf{H}^{t-1}_{j,:}) \quad \textrm{when} \quad (i,j) \in \mathcal{E}, $$ where $f$ is a single layer neural network and $\mathbf{A}^t_{ij}$ is the adjacency matrix generated by attention in layer $t$.

\subsubsection{Graph Autoencoders}
Graph autoencoders \cite{simonovsky2018graphvae} aim to learn low dimensional representation for an entire graphs and then reconstruct the graph via a decoder. These methods often stack multiple layers of ConvGNNs to model the encoder and decoder parts. In particular, the encoder parameters are optimized to fit $$p(\mathbf{z}|G,\mathbf{X}),$$ while the decoder attempts to recreate the original graph from the hidden embedding $$p(\hat{G}, \hat{\mathbf{X}}|\mathbf{z}).$$ Variational autoencoders exhibit the possibility to create variants of graphs by reconstructing modified $\hat{\mathbf{z}} = \mathbf{z} + \epsilon$ hidden states.

\subsubsection{Spatio-Temporal GNNs}
Spatio-temporal GNNs \cite{Bing2018stgcn} consider a sequence of graphs as input. Usually the graph topology $G$ remains constant, while node and edge features change over time. The key idea of Spatio-temporal GNNs is to consider spatial relationships (encoded via graph toplogy) and time relationships (change of features over time) simultaneously. Spatial encoding is performed via ConvGNNs, while recurrent memory cells (LSTM, GRU) are used to create hidden embedding along the time axis.

\section{Benchmarks}
\label{sec:benchmarks}
We introduce three different datasets for benchmarking node-level GNNs: Pubmed, Citeseer and Cora \cite{lu2003link, mccallum2000automating} are citation networks, in which nodes represent documents and edges are citation links. Node features correspond to word occurrences in the document's abstract and are encoded as bag-of-word vectors. Each document belongs to one or more classes and the task of the GNN is to predict a correct label for each of the target documents. Table~\ref{tab:datasets} shows dataset statistics.
\begin{table}[h]
  \centering        
  \begin{tabular}{lllll}
  \toprule
  Dataset & Nodes & Edges & Features & Classes \\ 
  \midrule
  Pubmed & 19,717 & 44,338 & 500 & 3 \\
  Citeseer & 3,327 & 4,732 & 3,703 & 6 \\
  Cora & 2,708 & 5,429 & 1,433 & 7 \\
  \bottomrule
  \end{tabular}
  \caption{\label{tab:datasets} Dataset statistics for Pubmed, Citeseer and Cora \cite{yang2016revisiting}.}
\end{table}

The most widely used train/test split is the one proposed by \cite{yang2016revisiting}. However, it is not clear if all the benchmarked methods follow the proposed train/test split rules or perform cross-validation. In general, all methods listed are trained on less than \SI{5}{\percent} labeled nodes. Table~\ref{tab:benchmarks} compares the classification accuracy on all three datasets introduced above.
\begin{table}[h]
  \centering      
  \scalebox{0.85}{  
    \begin{tabular}{llccc}
    \toprule
    Name & Type & Pubmed & Citeseer & Cora \\ 
    \midrule
    Planetoid \cite{yang2016revisiting} & Spatial & 77.2 & 64.7 & 75.7 \\
    GCN \cite{kipf2017graph} & Spatial & 79 & 70.3 & 81.5 \\
    GraphSAGE \cite{hamilton2017inductive} & Spatial & 78.3 & 71.1 & 83.3 \\
    Spectral CNN \cite{bruna13spectral} & Spectral & 73.9 & 58.9 & 73.3 \\
    ChebyNet \cite{defferrard2016convolutional} & Spectral & 74.4 & 69.8 & 81.2 \\
    Truncated Krylov \cite{luan2019break} & Spectral & 80.1 & 74.2 & 83.5 \\
    \bottomrule
    \end{tabular}
  }
  \caption{\label{tab:benchmarks} Classification accuracy on Pubmed, Citeseer and Cora for selected methods. Statistics taken from individual papers.}
\end{table}

\section{Applications}
\label{sec:applications}
Since graphs are a natural choice to represent data in many domains, GNNs have a variety of applications. Some of them are presented in this section. Many more examples can be found in \cite{zhou2018graph}.

\begin{figure}[tp]
  \centering
  \includegraphics[width=0.95\hsize]{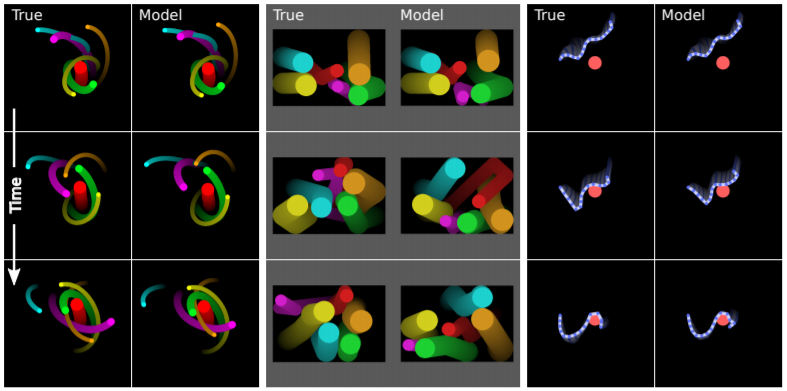}
  \caption{\label{fig:interactionet} Interaction networks predicting physical system states. 'True' is the ground truth, 'Model' is the prediction of the method initialized with state shown in first row. Image taken from \cite{battaglia2016interaction}.}
\end{figure}

\begin{figure}[tp]
  \centering
  \includegraphics[width=0.98\hsize]{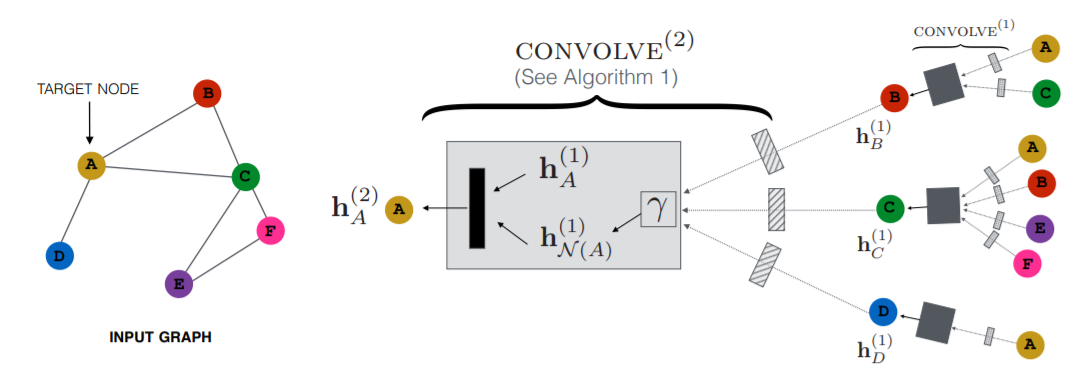}
  \caption{\label{fig:pinsage1} PinSage algorithm overview. In order to generate hidden embeddings for node A, a two layer ConvGNN is used to transform input values to hidden embeddings. Image taken from \cite{ying2018graph}.}
\end{figure}

\begin{figure}[tp]
  \centering
  \includegraphics[width=0.95\hsize]{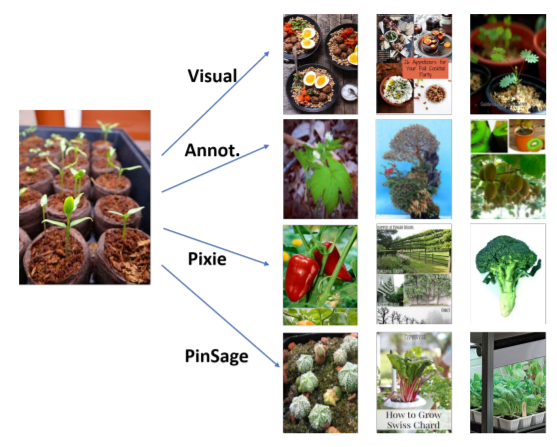}
  \caption{\label{fig:pinsage2} PinSage recommendation compared to other algorithms. Given an input image (left), PinSage recommends multiple images based on nearest neighbor queries on hidden embeddings created by a ConvGNN. Image taken from \cite{ying2018graph}.}
\end{figure}

\begin{figure}[tp]
  \centering
  \includegraphics[width=0.95\hsize]{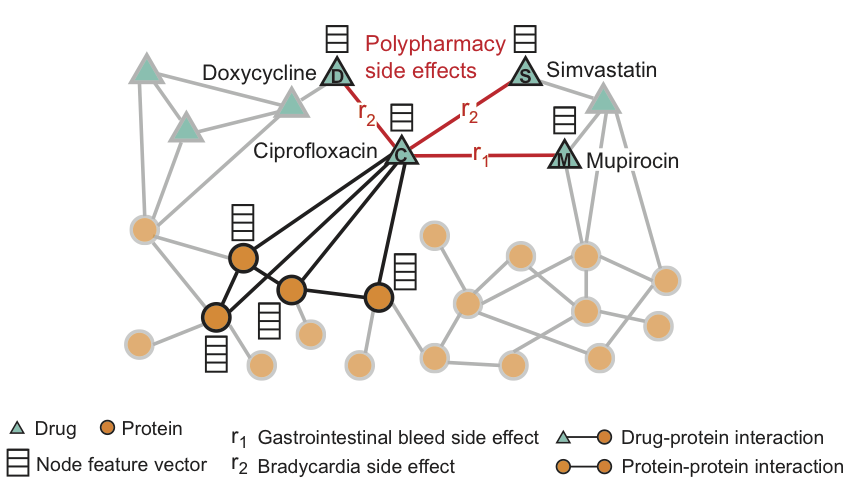}
  \caption{\label{fig:drugsideeffects} The polypharmacy graph models protein-protein, drug-protein, and drug-drug interactions. The network takes the former two as given and predicts edge-level output information for drug-drug interactions interpreted as the probability of side effects. Image taken from \cite{zitnik2018}.}
\end{figure}

\begin{figure}[tp]
  \centering
  \includegraphics[width=0.95\hsize]{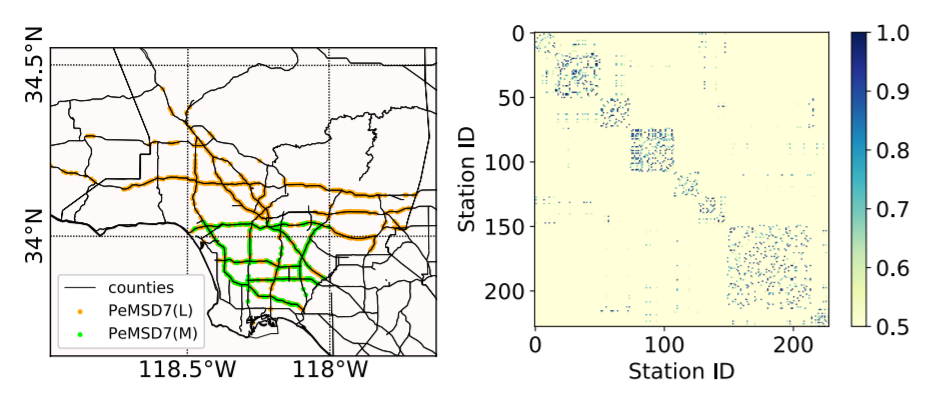}
  \caption{\label{fig:traffic} Left: roads with traffic sensor locations superimposed. Right, adjacency matrix of more than 200 road traffic sensors.}
\end{figure}

\begin{description}
    \item[Interaction Networks] \cite{battaglia2016interaction} propose a learnable physics engine  that considers the relation of objects and system state at time $t$ and then predicts the state for future timepoints. The input state is modelled as graph and the engine consists of a ConvGNN. See Figure~\ref{fig:interactionet} for a demonstration.
    \item[PinSage] \cite{ying2018graph} describe a large-scale graph convolution network for recommending pins in PInterest. The model generates hidden embeddings for all nodes (pins) given their relation to other nodes and input images. Figure~\ref{fig:pinsage1} describes the process of generating embeddings. The embeddings are then used for recommendation purposes using simple nearest neighbor queries as shown in Figure~\ref{fig:pinsage2}.
    \item[Drug Side Effects] \cite{zitnik2018} propose a ConvGNN to predict polypharmacy side effects based on drug and protein interaction. Simultaneous use of multiple drugs increases the risk of side-effects for the patient. Their network models protein-protein interactions, drug-protein interactions and predicts drug-drug interactions (side-effects). Figure~\ref{fig:drugsideeffects} illustrates the polypharmacy graph.
    \item[Traffic Forecasting] \cite{Bing2018stgcn} proposes a spatio-temporal ConvGNN for timely accurate traffic forecasting. The spatial axis comprises sensors measuring traffic at specific locations (nodes). The graph edges represent distances between network sensors. The traffic measurements of sensors are the features that change with time. Figure~\ref{fig:traffic} shows the placement of traffic sensors on roads across California. Given an input sequence, the model predicts future road traffic for each sensor location.
\end{description}

\section{Discussion}
\label{sec:discussion}
As illustrated in the last section, GNNs have shown great performance for prediction tasks in medicine, physics and social sciences. However, GNNs still pose open research questions.

\begin{description}
  \item[Over-smoothing] There is experimental evidence \cite{li2018deeper} that the performance of GNNs is inversely proportional to the stacking depth. At present time, shallow networks work better than deeply stacked networks. \cite{kipf2017graph, chen2019measuring} argue that the performance drop can be explained by an over-smoothing effect inherent to GNNs. This issue is reminiscent of contraction mapping issues of RecGNNs. The contraction mapping forces nearby nodes to create embeddings with decreasing distance. This leads to indistinguishable representations of
  nodes in different classes and impedes the performance of a GNN.
  \item[Scalability] ConvGNNs exploit local neighborhood graph information to generate node embeddings. In each layer information can propagate one hop further around the graph topology. Because of the over-smoothing issue, we cannot stack an arbitrary number of layers and still remain discriminative. Inescapably, information cannot propagate throughout the entire graph and is lost. The global filters of Spectral ConvGNNs are not exposed to this issue, but due to the eigen-decomposition they are limited to moderate graph sizes.
  \item[Heterogenous Graphs] Many GNNs are applied to homogenous graphs. Heterogenous graphs may consist of node types with varying semantics (e.g. factor-graphs) and it is still unclear how to handle those. For this reason, a practitioner in the field of GNNs will find it hard to choose a template for modelling a specific graph-related problem.
\end{description} 

\section{Conclusion}
\label{sec:conclusion}
In this work we introduced Graph Neural Networks (GNNs) to tackle problems in which graph structures are the natural way to represent data. In particular, we focused on Convolutional Graph Neural Networks (ConvGNN). We motivated these by the inspiring properties of Convolutions in the Euclidean domain. We showed that GNNS are capable of addressing many important machine learning problems in a wide variety of domains, such as medicine, physics and social sciences. 

We also find that the success of GNNs is strongly linked to the modeling of the input graph. In contrast to the Euclidean domain, where the graph structure is often implicit (e.g. grid-like in images), the graph structure in the non-Euclidean domain is often designed manually. This holds opportunities and risks for the success of a GNN and partly explains why many different solutions for similar tasks are proposed.

\clearpage
\bibliographystyle{ACM-Reference-Format}
\bibliography{graphnn}

\end{document}